\documentclass[twocolumn,pdflatex,sn-nature]{sn-jnl} 

\usepackage{geometry}
\usepackage{graphicx}%
\usepackage{multirow}%
\usepackage{amsmath,amssymb,amsfonts}%
\usepackage{amsthm}%
\usepackage{mathrsfs}%
\usepackage[title]{appendix}%
\usepackage{xcolor}%
\usepackage{textcomp}%
\usepackage{manyfoot}%
\usepackage{booktabs}%
\usepackage{algorithm}%
\usepackage{algorithmicx}%
\usepackage{algpseudocode}%
\usepackage{listings}%
\usepackage{hyperref}
\usepackage[capitalize]{cleveref}
\crefname{section}{Sec.}{Secs.}
\Crefname{section}{Section}{Sections}
\Crefname{table}{Table}{Tables}
\crefname{table}{Table}{Tables}
\Crefname{figure}{Figure}{Figures}
\crefname{figure}{Fig.}{Figs.}
\Crefname{equation}{Equation}{Equations}
\crefname{equation}{Eq.}{Eqs.}
\crefname{algocf}{alg.}{algs.}
\Crefname{algocf}{Algorithm}{Algorithms}

\begin{document}

\title[Article Title]{Empowering Functional Neuroimaging: A Pre-trained Generative Framework for Unified Representation of Neural Signals}

\author[1]{\fnm{Weiheng} \sur{Yao}}

\author[1]{\fnm{Xuhang} \sur{Chen}}
\author*[1]{\fnm{Shuqiang} \sur{Wang}}

\affil*[1]{\orgdiv{Shenzhen Institute of Advanced Technology}, \orgname{Chinese Academy of Sciences}, \orgaddress{\city{Shenzhen}, \postcode{518055}, \state{Guangdong}, \country{China}}}

\abstract{
Multimodal functional neuroimaging enables systematic analysis of brain mechanisms and provides discriminative representations for brain-computer interface (BCI) decoding. However, its acquisition is constrained by high costs and feasibility limitations. Moreover, underrepresentation of specific groups undermines fairness of BCI decoding model. To address these challenges, we propose a unified representation framework for multimodal functional neuroimaging via generative artificial intelligence (AI). By mapping multimodal functional neuroimaging into a unified representation space, the proposed framework is capable of generating data for acquisition-constrained modalities and underrepresented groups. Experiments show that the framework can generate data consistent with real brain activity patterns, provide insights into brain mechanisms, and improve performance on downstream tasks. More importantly, it can enhance model fairness by augmenting data for underrepresented groups. Overall, the framework offers a new paradigm for decreasing the cost of acquiring multimodal functional neuroimages and enhancing the fairness of BCI decoding models.
}

\maketitle

\begin{figure*}
    \centering
    \includegraphics[width=\linewidth]{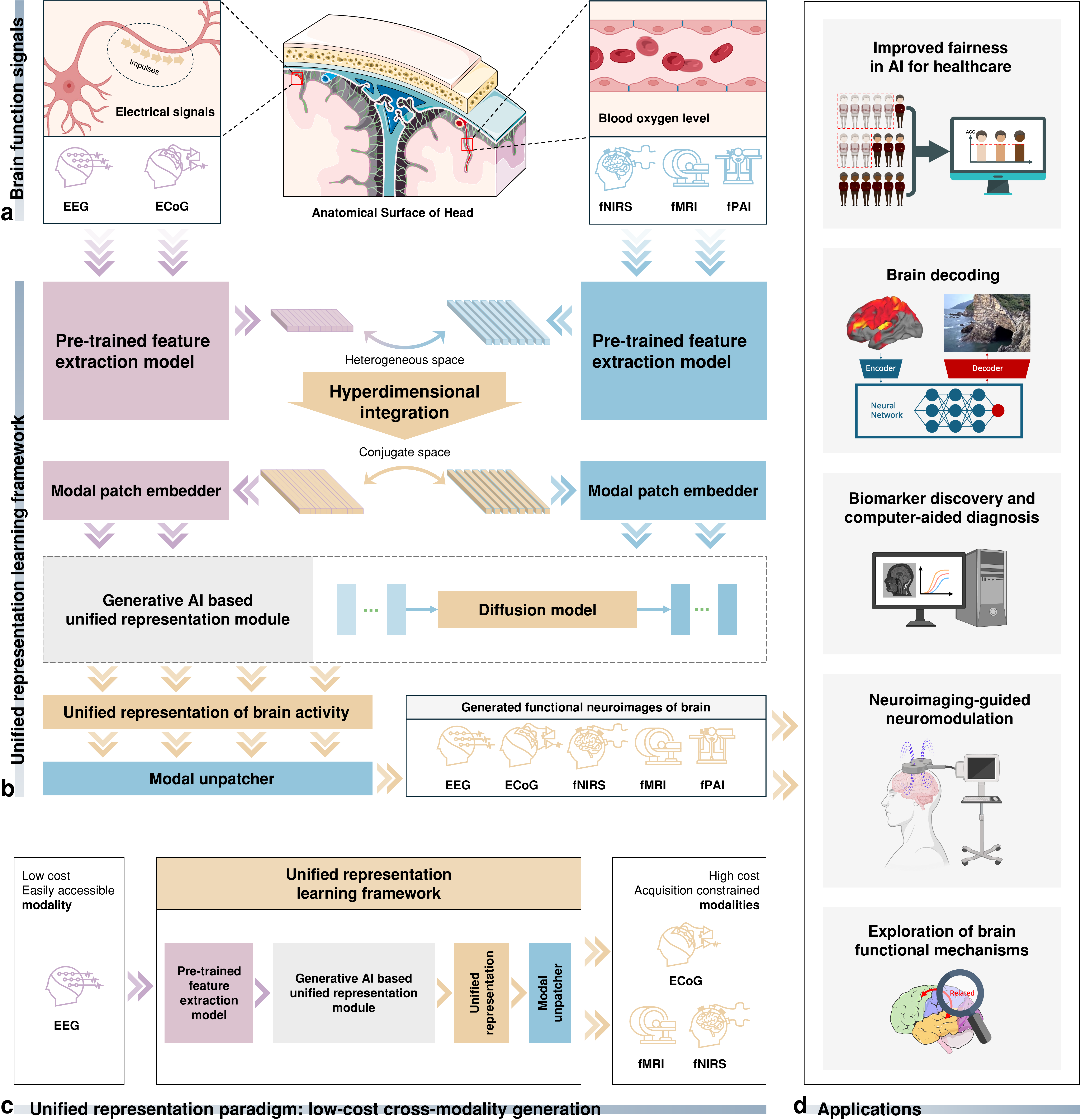}
    \caption{\textbf{Schematic of the proposed framework}.
    	\textbf{a,} Detection modalities for brain activity patterns, including EEG, ECoG, fNIRS, fMRI, and fPAI, along with their anatomical and physiological mechanism representations. The image labeled 'Anatomical Surface of Head' was adopted from Servier Medical Art by Servier (https://smart.servier.com) and modified by the authors under the following terms: CREATIVE COMMONS Attribution 3.0 Unported (CC BY 3.0).
    	\textbf{b,} Unified representation framework, comprising pre-trained feature extraction model, hyperdimensional integration, generative AI module, and modal unpatcher for reconstructing functional neuroimages.
    	\textbf{c,} The framework provides a new paradigm for cross-modal generation, transforming accessible, low-cost modalities (e.g., EEG) into high-cost neuroimaging signals (e.g., fMRI, ECoG, fNIRS), thereby bridging cost and accessibility gaps in brain imaging.
    	\textbf{d,} Applications of the proposed framework, including improved fairness in healthcare, BCI decoding, biomarker discovery, computer-aided diagnosis, neuroimaging-guided neuromodulation and exploration of brain functional mechanisms.}
    \label{fig1}
\end{figure*}

Multimodal functional neuroimaging is revolutionizing neuroscience by providing a comprehensive understanding of neural processes underlying cognition \cite{nature2}, emotion \cite{nmi1}, and behavior \cite{nmi2}. Each modality offers unique access to different aspects of brain activity. Electrophysiological techniques such as electroencephalography (EEG) directly measure neural electrical dynamics through scalp-recorded potentials, tracking rapid changes in brain activity. In contrast, electrocorticography (ECoG) provides superior spatial precision by placing electrodes directly on the cortical surface.
Functional magnetic resonance imaging (fMRI) remains the gold standard for non-invasive mapping of human brain function. It detects blood-oxygen-level-dependent (BOLD) signal fluctuations that indirectly reflect brain activity through associated metabolic demands. fMRI also provides whole-brain coverage, including deep cortical and subcortical structures\cite{natrevphy}.
In parallel, emerging optical and optoacoustic technologies such as functional near-infrared spectroscopy (fNIRS) and functional photoacoustic imaging (fPAI) \cite{new_fPAI_ref_placeholder} can also monitor cerebral haemodynamic responses.
By integrating these complementary modalities (\cref{fig1}(a)), researchers can achieve a more comprehensive characterization of brain mechanisms, with significant potential for BCI \cite{natureBCI} and neurovascular coupling research \cite{natureBOLD}.

Despite its transformative potential, the broad and equitable application of multimodal functional neuroimaging faces significant challenges. One major obstacle is the high cost and feasibility limitations associated with data acquisition. This is particularly evident for advanced modalities like fMRI, which require substantial financial investment, participant immobility during scanning, expert technical support, sophisticated acquisition protocols, and complex analysis pipelines \cite{naturefMRI, jing1}.
Exacerbating these accessibility challenges, available datasets often exhibit inherent biases. These biases primarily stem from the underrepresentation of specific groups. Institutions with limited financial resources, which serve these underrepresented groups, are less likely to contribute data from costly modalities like fMRI.
As a result, such data are frequently absent from large-scale repositories used to train AI models for BCI decoding and other applications \cite{supp_ref2, supp_ref3}.
This lack of representation critically undermines the fairness of these models. Models trained on imbalanced datasets typically exhibit diminished performance when applied to underrepresented groups, leading to unreliable insights and risking the perpetuation of healthcare inequities \cite{fairness_ktena2024generative}.
Addressing the issues of acquisition cost, data accessibility, and equitable representation is therefore essential. Only by overcoming these barriers can the full potential of multimodal neuroimaging be realized in both neuroscience research and applications.

Emerging advances in AI \cite{us+1, us+5}, particularly the advent of powerful pre-trained foundation models and generative AI techniques \cite{natmethods1, natmethods2} like generative adversarial networks\cite{us+2, us+4, us+9} and diffusion models, offer unprecedented potential to address these fundamental challenges \cite{us+6, us+8}.
Pre-trained models show strong potential in learning meaningful representations from heterogeneous multimodal data. Although these data differ significantly in resolution, signal properties, and informational content \cite{nc1, nc2}, pre-trained models have demonstrated the ability to map them into a shared space \cite{GMAI_moor2023foundation, jing7}.
Meanwhile, generative AI offers a transformative approach to solving data scarcity and imbalance \cite{jing13}. These models can generate high-quality \cite{us+3,us+10}, condition-specific neural data \cite{us+7, us+11, us+12}, enabling the generation of data for acquisition-constrained modalities \cite{jing4}. More importantly, they also facilitate the generation of data for underrepresented groups to bolster the fairness and generalizability of BCI decoding models driven by multimodal functional neuroimages.

In this work, a unified representation framework (\cref{fig1}(b)) for multimodal functional neuroimaging is proposed, addressing key challenges of cost, feasibility, and model fairness. The proposed framework uses pre-trained feature extraction models to map EEG signals, as well as fMRI and fNIRS signals, into a unified representation space.
Precise cross-modal alignment between electrophysiological patterns is then achieved by hyperdimensional integration.
Within the aligned representation space, a diffusion-based generative model learns the joint neurophysiological distribution to generate unified representations.
These unified representations are subsequently decoded into modality-specific outputs via a modal unpatcher, enabling reconstruction of the target BOLD.
Performance and capabilities of the proposed framework were validated using kinds of multimodal datasets, including simultaneously acquired EEG-fMRI and EEG-fNIRS datasets.
It accurately generates target modalities from EEG inputs while preserving key neurophysiological characteristics, including regional activation patterns and neural oscillatory dynamics. The framework can also maintain generalizability across diverse groups and cross-modal tasks. Significantly, it enhances the fairness of BCI decoding models through targeted data augmentation for underrepresented groups.
Overall, this work provides a new paradigm (\cref{fig1}(c)) that enables the data-driven reconstruction of high-cost, restricted-access neuroimaging modalities—such as fMRI—from readily available, low-cost signals like EEG. This approach opens up new paths for extensive applications (\cref{fig1}(d)), including fairness enhancing in AI for healthcare, BCI decoding \cite{nmi9}, biomarker discovery, computer-aided diagnosis, neuroimaging-guided neuromodulation and exploration of brain functional mechanisms.

\section*{Results}

The unified representation framework was rigorously evaluated using EEG-fMRI and EEG-fNIRS datasets to assess its performance. Brain regions were consistently delineated across all regional analyses (Destrieux atlas \cite{destrieux2010automatic}). The consistency of brain activity patterns, both temporally and spatially, was verified, as shown in \cref{fig2}.
SHAP-based analysis \cite{SHAP} reveals that the framework interprets brain mechanisms across three aspects: regional activity, connectivity patterns, and neural oscillation dynamics (\cref{fig3}).
The framework demonstrated cross-modal and inter-subject generalization (\cref{fig4}) and improved accuracy in downstream tasks, including clinical decision support for parkinson’s disease (PD) and brain decoding (\cref{fig5}). Critically, data augmentation enhanced model fairness by mitigating fMRI/fNIRS dataset imbalances that could introduce bias due to underrepresentation of specific groups (\cref{tab:fair}).
Collectively, these results underscore the framework’s efficacy in integrating multimodal neuroimaging data and BCI applications, advancing neurovascular coupling mechanism research, and cost-effective diagnostics.

\subsection*{The Proposed Framework Generates Data Consistent with Real Brain Activity Patterns}

\begin{figure*}
    \centering
    \includegraphics[width=\linewidth]{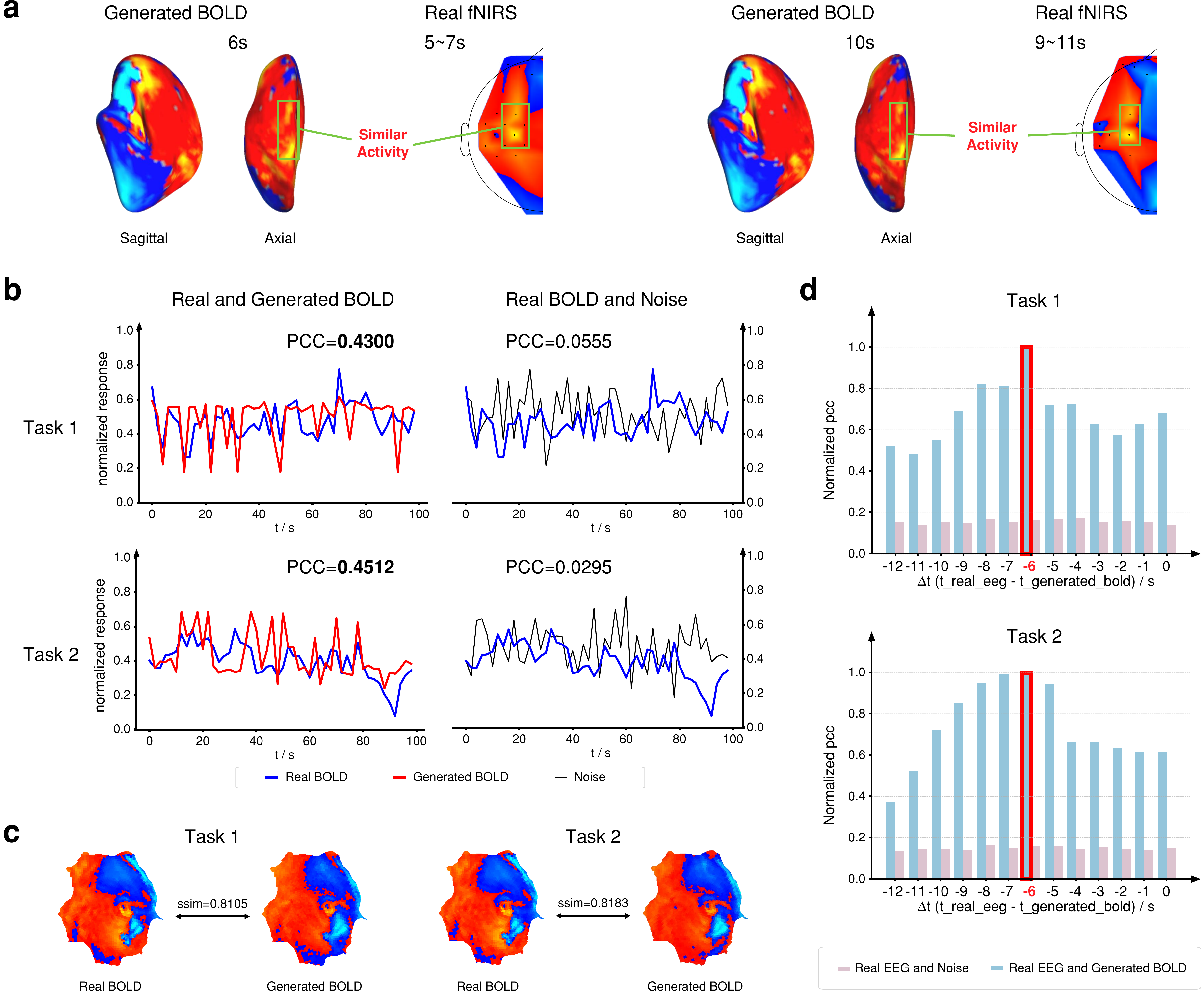}
    \caption{
\textbf{Consistency assessment of cross-modal generated BOLD and real signals.}
\textbf{a,} Cross-modal unified representation analysis: A model trained on EEG-fMRI data generates BOLD from EEG inputs in a simultaneously acquired EEG-fNIRS dataset. The generated BOLD exhibit activation patterns that closely correspond to those of real fNIRS.
\textbf{b,} Temporal consistency analysis: Comparison of real BOLD (blue) and generated BOLD (red) for Task 1 and Task 2, with noise (black) as a baseline. PCC between generated and real signals are significantly higher than those between noise and real signals, confirming the model's ability to capture temporal BOLD dynamics.
\textbf{c,}  Spatial consistency analysis: SSIM values for Task 1 (0.8105) and Task 2 (0.8183) demonstrate spatial consistency between real and generated BOLD.
\textbf{d,}  Temporal relationship analysis: Normalized PCC is computed between generated BOLD and real EEG inputs across time lags. For both Task 1 and Task 2, the correlation peaked at -6 seconds. This aligns with the known hemodynamic delay.
    }
    \label{fig2}
\end{figure*}

The capacity of the proposed framework to generate BOLD was evaluated using EEG-fMRI and EEG-fNIRS datasets, with a focus on its ability to unify multimodal representations while maintaining temporal and spatial consistency with real neuroimaging signals.

\textbf{Unified Representation.}
In this experiment, the framework employed a model trained on EEG-fMRI data to generate BOLD from EEG inputs in the EEG-fNIRS dataset. As shown in \cref{fig2}(a), the generated BOLD were directly compared with real fNIRS at corresponding time points. The results revealed similarity between the two modalities, particularly in motor-related brain regions. Both the generated BOLD and real fNIRS exhibited significant activation in the following anatomical areas: G\_and\_S\_paracentral (Paracentral lobule and sulcus), G\_front\_sup (Superior frontal gyrus) and S\_precentral-sup-part (Superior part of the precentral sulcus). This overlap highlights the framework's ability to preserve neurophysiologically meaningful features relevant to motor imagery tasks.

\textbf{Temporal Consistency.}
The temporal fidelity of the generated BOLD was assessed by computing Pearson correlation coefficients (PCC) between generated and real BOLD time courses (\cref{fig2}(b)). For Task 1, the PCC was 0.4300 (noise baseline: 0.0555); for Task 2, it was 0.4512 (noise baseline: 0.0295), both statistically significant. These results confirm the temporal consistency of the generated signals. Additionally, cross-modal temporal alignment between the generated BOLD and the corresponding EEG inputs was analyzed (\cref{fig2}(d)), revealing a hemodynamic delay of approximately 6 seconds. This result confirms that the model accurately captures the temporal relationship between electrophysiological and hemodynamic responses.

\textbf{Spatial Consistency.}
Spatial similarity between generated and real BOLD was quantified using the Structural Similarity Index Measure (SSIM). SSIM values of 0.8105 (Task 1) and 0.8183 (Task 2) (\cref{fig2}(c)) demonstrate strong spatial consistency, particularly in visual processing regions like G\_cuneus and G\_occipital\_inf.

Collectively, these results demonstrate that the framework is effective in integrating multimodal signals, preserving key activation patterns, and accurately capturing brain activity patterns across different imaging modalities.

\subsection*{The Unified Representations Provide Insights into Brain Mechanisms}

\begin{figure*}
    \centering
    \includegraphics[width=\linewidth]{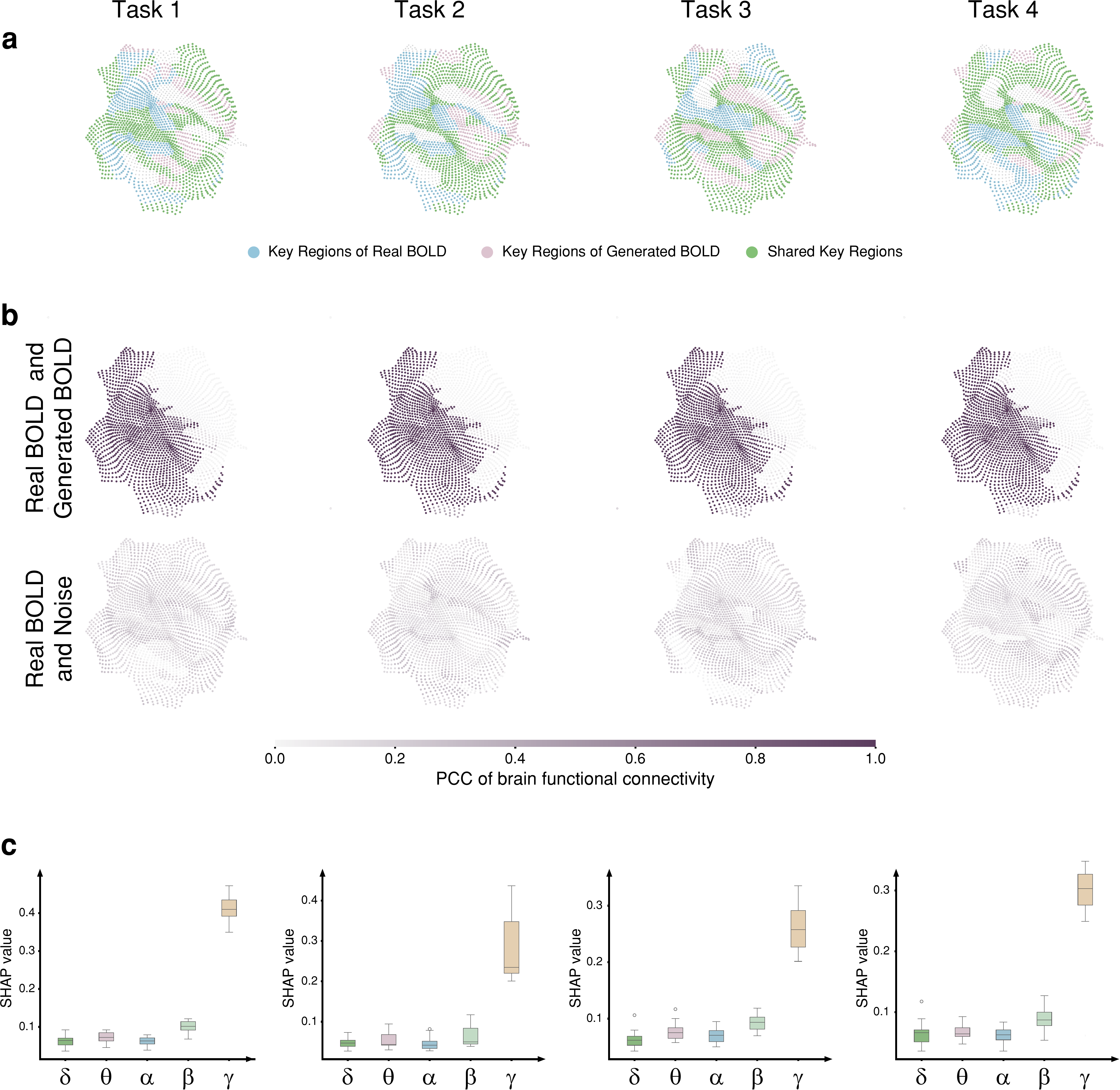}
    \caption{
    \textbf{Evaluation of unified representations in elucidating brain mechanisms.}
    \textbf{a,} Key brain regions identified via SHAP analysis for brain decoding. Green regions (largest area) indicate shared key areas between real and generated BOLD, while blue and pink regions are uniquely activated in real and generated signals.
    \textbf{b,} Functional connectivity correlation analysis: The correlation between the functional connectivity of generated BOLD and real BOLD is significantly higher than that between real BOLD and noise. This confirms that the generated BOLD capture meaningful brain functional connectivity patterns.
	\textbf{c,} The contribution of EEG frequency bands to the generated BOLD was evaluated using SHAP values. The gamma band consistently shows the highest contribution across all conditions. Each boxplot displays the median (central line), interquartile range (box, spanning the 25th to 75th percentiles), and whiskers (extending to the minimum and maximum SHAP values within 1.5 times the interquartile range from the quartiles), with outliers shown as individual points beyond the whiskers.
    }
    \label{fig3}
\end{figure*}

The interpretability of brain mechanisms underlying the data generated by the proposed framework was assessed using multimodal data from naturalistic visual tasks (Task 1, Task 2, Task 3, Task 4), with an emphasis on their consistency with established neurophysiological principles.

\textbf{Key Brain Regions.}
SHAP analysis (\cref{fig3}(a)) identified key brain regions contributing to neural activity predictions based on both real and generated BOLD across all tasks. A substantial overlap (mean: 54.64 $\pm$ 1.34\%, $n=22$) was observed between the two modalities, with shared activation in G\_cuneus (primary visual cortex), G\_temporal\_sup (superior temporal gyrus), and G\_precuneus—regions critical for visual processing. Additionally, regions uniquely activated in the generated signals (15.93\%) included S\_parieto\_occipital, while real BOLD engaged subcortical structures more prominently (29.44\%), such as the thalamus.

\textbf{Functional Connectivity.}
Functional connectivity patterns derived from the generated and real BOLD were compared across the four visual tasks (\cref{fig3}(b)). The real BOLD showed significantly greater similarity to the generated BOLD than to the noise baseline. Specifically, PCC values for the generated signals were consistently higher in most visual-processing-related regions, whereas the noise baseline yielded predominantly low PCC values, indicating minimal functional correspondence with the real BOLD. These results demonstrate that the framework effectively preserves neurophysiologically meaningful connectivity patterns.

\textbf{EEG Frequency Contributions.}
SHAP analysis (\cref{fig3}(c)) evaluated the contributions of different EEG frequency bands to the generated BOLD across all tasks. The gamma band consistently showed the highest SHAP values, significantly exceeding those of the beta, alpha, theta, and delta bands. This dominance of the gamma band aligns with its well-established role in high-level cognitive processes such as visual attention and perception. In contrast, the relatively lower contributions from alpha, theta, and delta bands may reflect their association with more diffuse or resting-state neural activity, which is less directly linked to the task-specific visual processing demands.

Collectively, these findings confirm that the unified representation framework captures neurophysiologically interpretable brain mechanisms, demonstrating its ability to learn a coherent and functionally relevant mapping across modalities.

\subsection*{The Proposed Framework Enables Cross-Modal and Cross-Subject Generalization}\label{results-cross}

\begin{figure*}
    \centering
    \includegraphics[width=\linewidth]{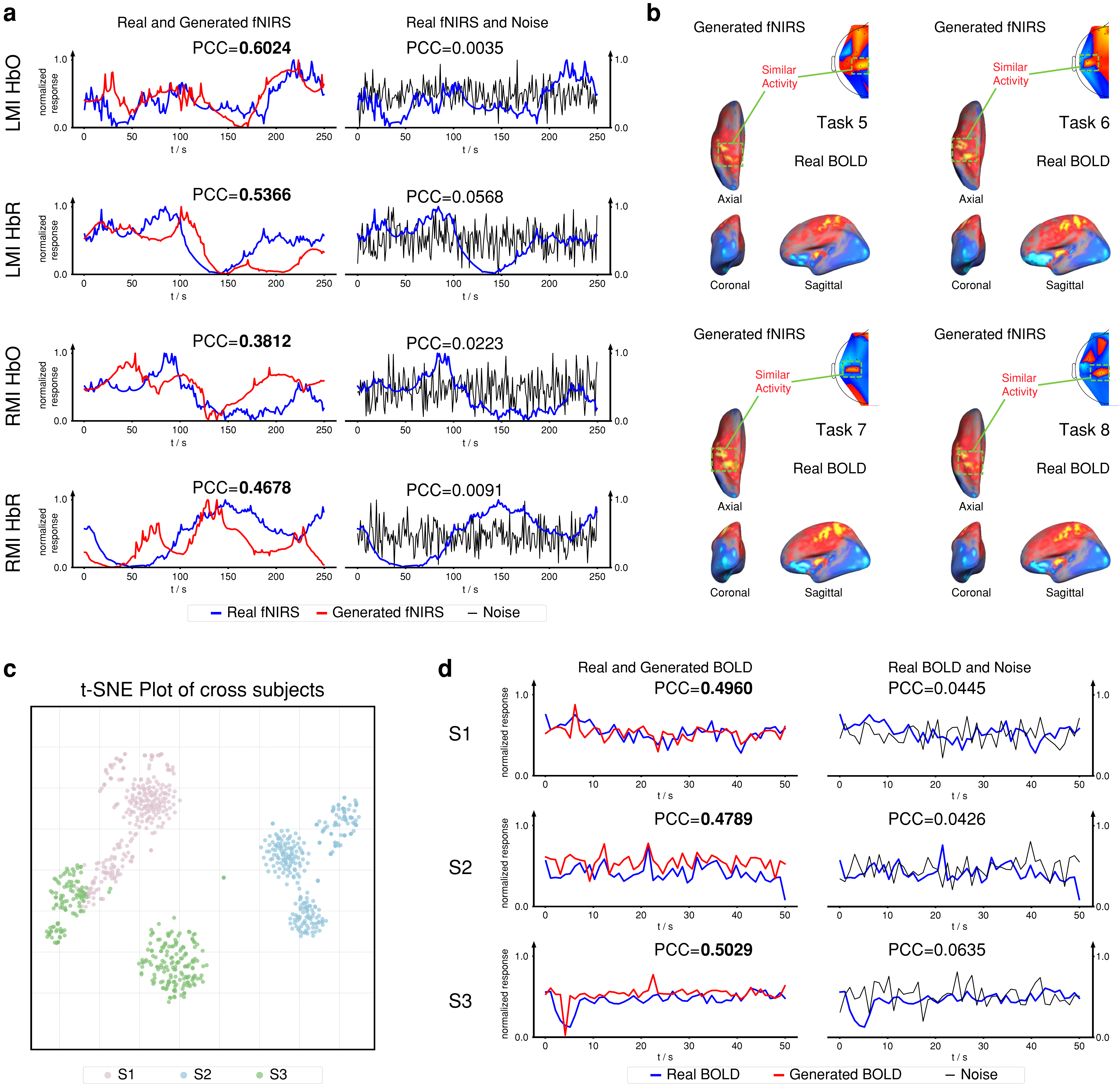}
    \caption{
\textbf{Assessment of generalization in the proposed framework.}
\textbf{a,} Comparison of generated fNIRS (red) with real fNIRS (blue) for left-hand motor imagery (LMI) and right-hand motor imagery (RMI). PCC between generated and real fNIRS are significantly higher than those between noise (black) and real fNIRS, confirming the framework's effectiveness in cross-modal generation across diverse neuroimaging modalities.
\textbf{b,} Cross-modal generalization: A model trained on EEG-fNIRS dataset generates fNIRS from EEG inputs within a simultaneously acquired EEG-fMRI dataset. The generated fNIRS exhibit activation patterns comparable to real BOLD across visual stimulus tasks, demonstrating cross-modal generalization.
\textbf{c,} t-SNE visualization of generated BOLD across subjects. Distinct clustering by individual subject indicates that the framework preserves subject-specific neural features.
\textbf{d,} Cross-subject similarity analysis: PCC values between generated BOLD (red) and real BOLD (blue) substantially exceed those between noise (black) and real BOLD, confirming both the consistency and subject-specific revelance of the generated outputs.
    }
    \label{fig4}
\end{figure*}

The performance of the unified representation framework was evaluated using multimodal data from visual tasks (Task 1, Task 2, Task 3, Task 4) and motor imagery states (left and right hand), with a focus on assessing its generalizability through various cross-modal and cross-subject analyses.

\textbf{fNIRS Generation from EEG.}
In this experiment, the framework generated fNIRS from EEG inputs, focusing on oxy-hemoglobin (HbO) and deoxy-hemoglobin (HbR) concentrations in motor imagery-related regions (\cref{fig4}(a)). PCC between the generated and real fNIRS were as follows: 0.6024 (LMI HbO), 0.5366 (LMI HbR), 0.3812 (RMI HbO), and 0.4678 (RMI HbR)—all significantly higher than the noise baseline (0.0229 $\pm$ 0.0120).

\textbf{Activation Pattern Consistency Between Generated fNIRS and BOLD.}
The activation patterns of the generated fNIRS signals were compared with those of real BOLD across the four visual tasks (\cref{fig4}(b)). High similarity was observed in motor-related brain regions, including the G\_precentral (precentral gyrus), S\_central (central sulcus), G\_postcentral (postcentral gyrus), S\_precentral-inf-part (inferior precentral sulcus), and S\_precentral-sup-part (superior precentral sulcus). Both modalities exhibited prominent activation in these areas, similar to the findings presented in \cref{fig2}(a).

\textbf{t-SNE Analysis for Subject Differentiation.}
t-SNE analysis (\cref{fig4}(c)) conducted on three subjects from the test set (not included in training) revealed distinct clustering patterns. This result indicates that the framework effectively captures subject-specific features, enabling differentiation among individuals.

\textbf{Cross-Subject Signal Consistency.}
Cross-subject consistency was assessed by computing PCC values between generated and real BOLD for three independent subjects (\cref{fig4}(d)), yielding correlations of 0.4960 (sub-1), 0.4789 (sub-2), and 0.5029 (sub-3). These values substantially exceeded the noise baseline (0.0502 $\pm$ 0.0133), demonstrating an approximately 9-fold improvement in signal fidelity.

Collectively, these results demonstrate the framework's strong generalization across modalities and subjects, integrating EEG, fNIRS, and fMRI data while maintaining cross-modal and cross-individual consistency.

\subsection*{The Unified Representation Improves Performance on Downstream Tasks}

\begin{figure*}
    \centering
    \includegraphics[width=\linewidth]{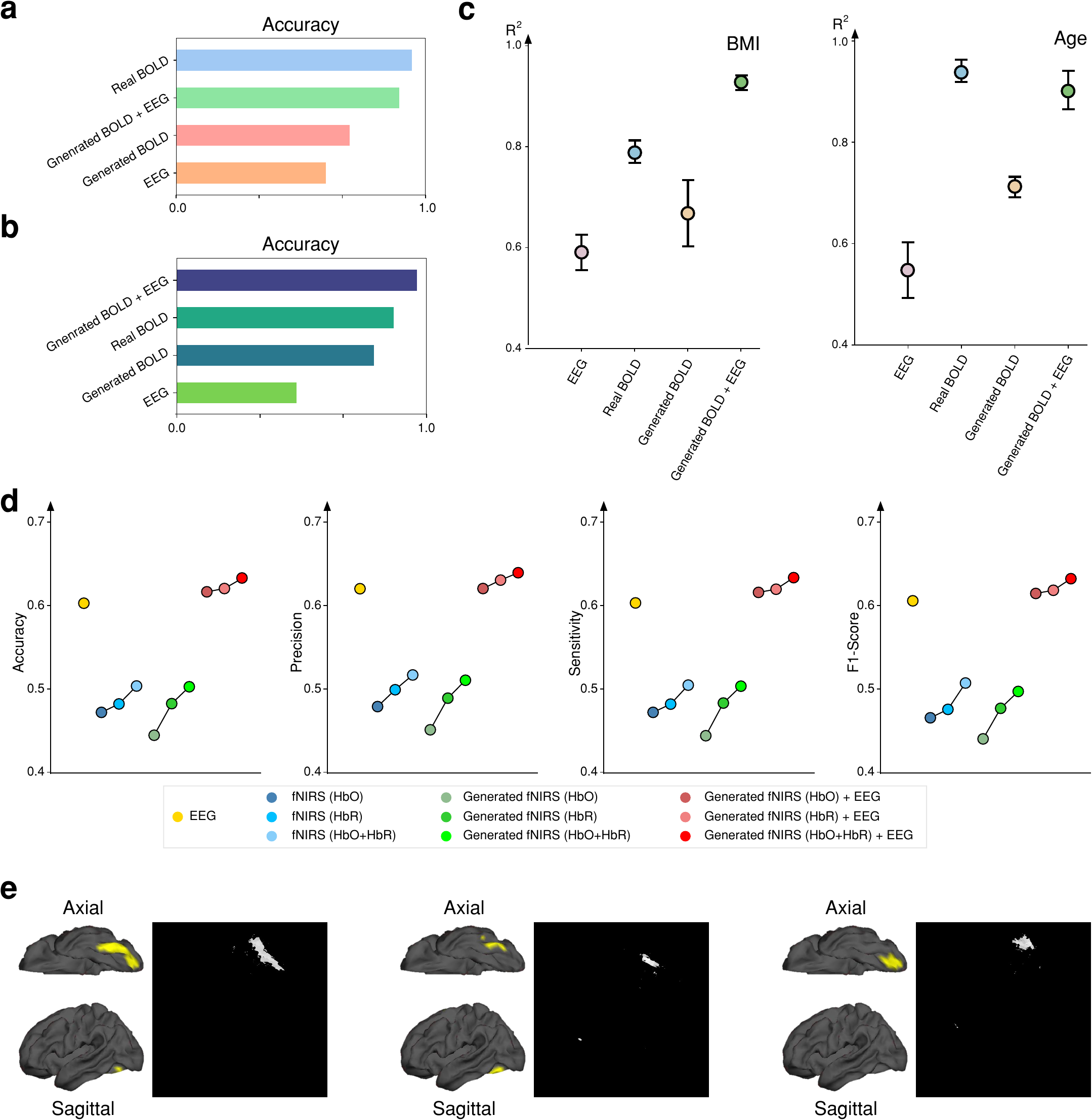}
    \caption{
\textbf{Evaluation of unified representations for brain decoding and clinical applications.}
\textbf{a,} Within-subject visual stimulus decoding: Generated BOLD, when combined with EEG, outperform EEG alone and achieve performance comparable to that of real BOLD, demonstrating their effectiveness in enhancing decoding accuracy.
\textbf{b,} Cross-subject visual stimulus decoding: Simliar to the within-subject results, the combination of generated BOLD and EEG significantly outperforms EEG alone and closely matches the performance of real BOLD.
\textbf{c,} Cross-subject continuous value prediction for physiological indicators: The integration of generated BOLD with EEG achieved superior performance compared to EEG alone.
\textbf{d,} Motor imagery state prediction using unified representation of EEG and fNIRS: Combining generated fNIRS with EEG achieves the best overall performance across all evaluation metrics. HbR outperforms HbO, consistent with its greater sensitivity to neural activity, while the HbR–HbO combination delivers higher prediction accuracy.
\textbf{e,} Comparison of BOLD difference maps generated from EEG in healthy subjects and PD patients: Abnormal regions are visualized across two orthogonal brain planes (sagittal and axial), with yellow areas indicating significant group-level differences. All three sets of results, derived from distinct subject group, identified the similar abnormal brain regions.
    }
    \label{fig5}
\end{figure*}

The practical utility of the unified representation framework was evaluated through a range of downstream tasks, including visual stimulus decoding, continuous value prediction, motor imagery state prediction, and BOLD difference mapping for PD (\cref{fig5}).

\textbf{Visual Stimulus Decoding.}
In both within-subject (\cref{fig5}(a)) and cross-subject (\cref{fig5}(b)) settings, the integration of generated BOLD with EEG consistently outperformed EEG alone. Prediction accuracies were within 5\% of those achieved using real BOLD in within-subject scenarios and improved by up to 10\% in cross-subject settings.

\textbf{Continuous Value Prediction.}
In cross-subject EEG-fMRI settings (\cref{fig5}(c)), the combination of generated BOLD and EEG demonstrated predictive performance for both BMI ($R^2 = 0.93$) and age ($R^2 = 0.90$). Notably, the BMI prediction accuracy exceeded not only EEG alone ($R^2 = 0.59$) but also real BOLD ($R^2 = 0.79$), suggesting that the generated BOLD may capture additional neurostructural information relevant to body mass index.

\textbf{Motor Imagery State Prediction.}
With the combination of EEG and generated fNIRS (\cref{fig5}(d)), the framework achieved a mean classification accuracy of 0.63 for motor imagery state prediction tasks—superior to that of EEG alone (0.60), as well as fNIRS(HbO) (0.47) or fNIRS(HbR) (0.48) alone. Notably, prediction based on fNIRS(HbR) $+$ EEG (0.62) slightly outperformed that based on fNIRS(HbO) $+$ EEG(0.61). This differential performance likely reflects the closer coupling with neural activity of fNIRS(HbR).

\textbf{Clinical Decision Support Potential.}
BOLD difference maps (\cref{fig5}(e)) were computed based on BOLD generated from EEG of both healthy control group and PD patients. These maps revealed significant deviations in several brain regions, including S\_oc-temp\_med\_and\_Lingual (Medial occipito-temporal sulcus and lingual sulcus), G\_oc-temp\_lat-fusifor (Lateral occipito-temporal gyrus) and G\_and\_S\_occipital\_inf (Inferior occipital gyrus and sulcus). In PD, these brain regions exhibit significant structural changes. Notably, in patients with visual hallucinations, the degree of atrophy correlates with hallucination severityp \cite{nature+2}.

Collectively, these results affirm the framework's strong applicability to diverse downstream tasks, achieving performance levels close to those obtained with original imaging modalities. By enabling effective multimodal integration, the framework can reduce operational costs by an estimated 90\%, primarily due to the cost disparity between fMRI and EEG. This substantial reduction enhances the accessibility of advanced neuroimaging techniques for clinical applications.

\subsection*{Cross-Modal Data Augmentation Enhances Fairness of BCI Decoding Models}

To evaluate the capability of the unified representation framework in cross-modal data generation and on enhancing fairness, experiments were conducted on both EEG-fMRI and EEG-fNIRS datasets. These experiments aimed to address fairness deficits stemming from imbalanced data distributions arising from underrepresented specific groups or conditions—a prevalent challenge in neuroimaging studies. By leveraging the generative capacity of proposed pre-trained framework, data were generated to augment underrepresented classes, and downstream task performance was assessed, focusing on predictive accuracy, stability, and fairness.

In the fMRI dataset, neural responses from four visual tasks (Task 5, Task 6, Task 7, Task 8)were used to predict task categories, with performance evaluated using F1-score under five-fold cross-validation. Data imbalance was simulated by reducing the sample size of one task to 30 (compared to 150 for others), and this process was repeated for each task. The framework then generated 120 samples per underrepresented class to rebalance all classes to 150 samples. For the fNIRS dataset, motor imagery tasks (LMI and RMI) were analyzed to classify imagery type. One class was reduced to 30 samples (vs. 120 for the other), and three-fold cross-validation was applied. The framework generated 90 samples to restore balance. Performance was evaluated before and after data augmentation.

\begin{table*}[]
\centering
\caption{\textbf{Evaluation of model fairness improvements via the proposed framework.} Each row represents a group of distinct data configuration, where the left side of the table shows the sample composition across tasks, and the right side presents the performance in BCI decoding. Gen in the Sample Size column indicates the number of generated samples. }
\label{tab:fair}
\renewcommand\arraystretch{1.25}
\resizebox{\linewidth}{!}{%
\begin{tabular}{cccccccc}
\toprule
\multicolumn{8}{c}{\textbf{Improving Fairness in Visual Stimulus Decoding}} \\ \midrule
\multicolumn{4}{c|}{\textbf{Sample Size}} &
  \multicolumn{4}{c}{\textbf{F1-Score}} \\ \midrule
\textbf{Task 5} &
  \textbf{Task 6} &
  \textbf{Task 7} &
  \multicolumn{1}{c|}{\textbf{Task 8}} &
  \textbf{Task 5} &
  \textbf{Task 6} &
  \textbf{Task 7} &
  \textbf{Task 8} \\ \midrule
\textbf{30} &
  150 &
  150 &
  \multicolumn{1}{c|}{150} &
  {\color[HTML]{000000} \textbf{0.539 $\pm$ 0.397}} &
  0.867 $\pm$ 0.166 &
  0.879 $\pm$ 0.085 &
  0.879 $\pm$ 0.089 \\
\textbf{30 + 120(Gen)} &
  150 &
  150 &
  \multicolumn{1}{c|}{150} &
  {\color[HTML]{000000} \textbf{0.881 $\pm$ 0.040}} &
  0.796 $\pm$ 0.181 &
  0.749 $\pm$ 0.115 &
  0.774 $\pm$ 0.132 \\ \hline
150 &
  \textbf{30} &
  150 &
  \multicolumn{1}{c|}{150} &
  0.864 $\pm$ 0.140 &
  {\color[HTML]{000000} \textbf{0.478 $\pm$ 0.347}} &
  0.782 $\pm$ 0.101 &
  0.840 $\pm$ 0.038 \\
150 &
  \textbf{30 + 120(Gen)} &
  150 &
  \multicolumn{1}{c|}{150} &
  0.737 $\pm$ 0.253 &
  {\color[HTML]{000000} \textbf{0.931 $\pm$ 0.033}} &
  0.706 $\pm$ 0.164 &
  0.878 $\pm$ 0.065 \\ \hline
150 &
  150 &
  \textbf{30} &
  \multicolumn{1}{c|}{150} &
  0.847 $\pm$ 0.159 &
  0.825 $\pm$ 0.169 &
  {\color[HTML]{000000} \textbf{0.522 $\pm$ 0.325}} &
  {\color[HTML]{000000} 0.845 $\pm$ 0.121} \\
150 &
  150 &
  \textbf{30 + 120(Gen)} &
  \multicolumn{1}{c|}{150} &
  0.819 $\pm$ 0.142 &
  0.771 $\pm$ 0.160 &
  {\color[HTML]{000000} \textbf{0.923 $\pm$ 0.022}} &
  {\color[HTML]{000000} 0.777 $\pm$ 0.083} \\ \hline
150 &
  150 &
  150 &
  \multicolumn{1}{c|}{\textbf{30}} &
  0.915 $\pm$ 0.031 &
  0.949 $\pm$ 0.035 &
  {\color[HTML]{000000} 0.883 $\pm$ 0.025} &
  {\color[HTML]{000000} \textbf{0.686 $\pm$ 0.201}} \\
150 &
  150 &
  150 &
  \multicolumn{1}{c|}{\textbf{30 + 120(Gen)}} &
  0.928 $\pm$ 0.017 &
  0.875 $\pm$ 0.072 &
  {\color[HTML]{000000} 0.773 $\pm$ 0.110} &
  {\color[HTML]{000000} \textbf{0.935 $\pm$ 0.033}} \\ \bottomrule
  \multicolumn{8}{l}{} \\ \toprule
\multicolumn{8}{c}{\textbf{Improving Fairness in Motor Imagery State Prediction}} \\ \midrule
\multicolumn{4}{c|}{\textbf{Sample Size}} &
  \multicolumn{4}{c}{\textbf{F1-Score}} \\ \midrule
\multicolumn{2}{c}{\textbf{LMI}} &
  \multicolumn{2}{c|}{\textbf{RMI}} &
  \multicolumn{2}{c}{\textbf{LMI}} &
  \multicolumn{2}{c}{\textbf{RMI}} \\ \midrule
\multicolumn{2}{c}{\textbf{30}} &
  \multicolumn{2}{c|}{120} &
  \multicolumn{2}{c}{\textbf{0.037 $\pm$ 0.052}} &
  \multicolumn{2}{c}{0.840 $\pm$ 0.025} \\
\multicolumn{2}{c}{\textbf{30 + 90(Gen)}} &
  \multicolumn{2}{c|}{120} &
  \multicolumn{2}{c}{\textbf{0.743 $\pm$ 0.016}} &
  \multicolumn{2}{c}{0.800 $\pm$ 0.019} \\ \hline
\multicolumn{2}{c}{120} &
  \multicolumn{2}{c|}{\textbf{30}} &
  \multicolumn{2}{c}{0.860 $\pm$ 0.029} &
  \multicolumn{2}{c}{\textbf{0.086 $\pm$ 0.061}} \\
\multicolumn{2}{c}{120} &
  \multicolumn{2}{c|}{\textbf{30 + 90(Gen)}} &
  \multicolumn{2}{c}{0.781 $\pm$ 0.008} &
  \multicolumn{2}{c}{\textbf{0.737 $\pm$ 0.034}} \\ \bottomrule
\end{tabular}%
}
\end{table*}

The fairness of the unified representation framework under imbalanced conditions was quantitatively assessed, with results summarized in \cref{tab:fair}. In the fMRI dataset, underrepresented categories such as Task 5 exhibited significantly lower F1-scores (0.539 $\pm$ 0.397) and higher variance compared to majority classes. After data augmentation, the performance of these minority classes improved markedly—for example, Task 5 achieved an F1-score of 0.881 $\pm$ 0.040—approaching that of the majority classes, while the variance decreased substantially (from 0.397 to 0.040).
In the fNIRS dataset, for instance, the underrepresented LMI class (initially 30 samples) yielded a very low F1-score of 0.037 $\pm$ 0.052, whereas the well-represented RMI class achieved 0.840 $\pm$ 0.025. Following augmentation, the F1-score for LMI increased to 0.743 $\pm$ 0.016, while RMI slightly declined to 0.800 $\pm$ 0.019, resulting in a more balanced classification performance across classes.
These findings demonstrate that the proposed framework enhances fairness in neuroimaging applications through data augmentation. This approach effectively reduces performance disparities between underrepresented and well-represented groups, enhances prediction stability, and promotes equitable outcomes in multimodal brain signal analysis.

\section*{Discussion}
\label{sec12}

\textbf{Analysis of Experimental Results.} The experimental results validate the feasibility and transformative potential of the proposed unified representation framework for BCI decoding \cite{nbe1,nm2}. Temporal-spatial consistency analyses (\cref{fig2}, \cref{fig4}) demonstrate accurate reconstruction of functional neuroimaging data, preserving both spatial activation patterns and temporal dynamics. Interpretability analyses (\cref{fig3}) further confirm that the framework captures biologically plausible representations. The EEG gamma-band activity and functional connectivity patterns align well with established neurophysiological principles \cite{nature+1}. These findings establish the framework as a powerful tool for multimodal neuroimaging, effectively bridging low-cost EEG with high-fidelity imaging signals.

The generalization capability of proposed framework (\cref{fig4}) highlights its versatility in capturing individual variability while maintaining consistent performance across subjects, tasks, and modalities. In contrast to modality-specific models, this generalizability enables broader applicability across diverse datasets and multi-task settings. Downstream applications (\cref{fig5}) further underscore its practical utility, where the generated BOLD and fNIRS signals significantly enhance decoding performance in tasks such as visual stimulus decoding, motor imagery state prediction, and BCI related inference. These improvements suggest promising applications in BCI \cite{nbe2,nmi13}, biomarker discovery \cite{nmi11,nmi12}, and clinical diagnostics \cite{nm3,nc3}.
Notably, fairness analyses (\cref{tab:fair}) reveal that the cross-modal generative approach mitigates performance disparities arising from imbalanced datasets. For underrepresented tasks, F1-scores show substantial improvement (e.g., from 0.539 $\pm$ 0.397 to 0.881 $\pm$ 0.040 for Task 5), along with reduced variance. While there is a modest decline in performance on majority tasks (e.g., Task 6, from 0.949 $\pm$ 0.035 to 0.875 $\pm$ 0.072), this trade-off reflects a shift in data distribution due to data augmentation. Importantly, this compromise is ethically justified by the need for equitable representation—particularly for underserved clinical populations. By addressing data imbalances, the proposed framework contributes to advancing model fairness, a critical objective in global neuroscience research \cite{nbe+1}.

\textbf{Prospects of the Unified Representation Framework.} The proposed framework opens up transformative opportunities for neuroimaging and clinical translation. By leveraging low-cost EEG to generate fMRI and fNIRS signals—reducing equipment costs by up to 90\%—it significantly broadens access to advanced brain decoding, particularly in resource-limited settings \cite{nmi16,nmi17}. The diffusion-based generative model, which captures complex spatiotemporal dependencies, can be further enhanced through emerging AI architectures. For instance, variational autoencoders may improve computational efficiency, while transformer could better capture long-range temporal dependencies, thereby enhancing signal fidelity.

The extensibility of the framework supports its integration with a wide range of modalities, including magnetoencephalography (MEG) \cite{nc4}, positron emission tomography (PET) \cite{nmi14}, and novel non-invasive techniques \cite{nmi15}. While the framework demonstrates potential for synthesizing ECoG from EEG, as illustrated in the framework design (\cref{fig1}(c)), the current work focuses on the successful synthesis of fMRI and fNIRS from EEG, as presented in the results. The EEG-to-ECoG generations still in its preliminary stages due to challenges in data acquisition, and thus was not included in this work. This multimodal compatibility has the potential to combine high temporal resolution with metabolic and hemodynamic insights, redefining how brain function is represented. Such integration challenges traditional, modality-specific paradigms and supports a more holistic understanding of neural activity.

Looking ahead, fairness considerations further amplify the framework’s societal impact. By improving performance on underrepresented tasks, it helps mitigate biases that might otherwise exacerbate diagnostic disparities—especially for rare neurological conditions. Future refinements could focus on optimizing data generation to achieve a better balance between performance and equity. Additionally, integrating the framework with wearable EEG devices could enable real-time, naturalistic brain monitoring, paving the way for innovative applications in neurofeedback therapy and BCI.
These advancements position the framework as a foundational tool for scalable, equitable, and precise brain decoding, with far-reaching implications for personalized medicine and global health.

\textbf{Limitations.} Despite its promising capabilities, the proposed framework faces several limitations that warrant further investigation. A primary challenge is the scarcity of simultaneously acquired multimodal neuroimaging data—most available datasets include fewer than 30 subjects—which limits the model's ability to capture inter-individual variability, particularly in specialized populations such as those with neurological disorders. This scarcity of diverse and representative data is a systemic challenge in neuroimaging and limits the framework’s generative capacity. As illustrated in Extended Data \ref{fig:extended}, some failures in BOLD signal reconstruction show mismatch in critical activity in brain regions and time points, suggesting overfitting to dominant patterns due to insufficient paired EEG-fMRI data. Future work could address this limitation through unsupervised learning strategies or federated data pooling to enhance training data diversity without compromising privacy.

Another key limitation lies in the precision of cross-modal alignment. While the framework effectively models the broad physiological coupling between EEG and BOLD, it struggles to capture subtle variations arising from regional differences in brain activity or task-specific cognitive states. This shortcoming is attributed to individual neuroplasticity and the complex, context-dependent interactions between electrical and hemodynamic responses. Furthermore, the limited heterogeneity in the training data exacerbates this issue, restricting the framework’s generalizability across diverse conditions and populations. Addressing these challenges will require innovative data acquisition approaches, such as multi-site collaborative studies, as well as advanced modeling techniques—including time-frequency decomposition methods like wavelet transforms—to better capture dynamic neural processes and improve generalizability across heterogeneous tasks and subject groups.

\section*{Methods}\label{sec11}

This work proposes a novel unified representation framework designed to bridge the gap between cost-effective, widely accessible EEG signals and high-cost, less accessible neuroimaging modalities such as fMRI and fNIRS. The framework integrates pre-trained feature extraction, hyperdimensional alignment, and diffusion-based generative modeling to capture complex spatiotemporal brain dynamics. It achieves strong cross-modal and inter-subject generalization. This approach enables the translation of EEG signals into high-fidelity hemodynamic responses, supporting cost-effective neuroimaging analysis and facilitating downstream applications including clinical decision support for PD and BCI decoding.

The proposed framework employs pretrained feature extraction models to transform raw EEG and target modality (fMRI or fNIRS) data into high-dimensional feature representations. These features are then fused into a shared spatiotemporal conjugate domain through a hyperdimensional integration strategy that preserves anatomical and physiological relationships across modalities. A Diffusion Transformer (DiT) based generative module \cite{DiT_peebles2022scalable} further refines this unified representation by learning latent temporal and spatial dependencies between EEG and the target imaging modalities during training. During inference, the trained model generates high-fidelity fMRI or fNIRS signals solely from EEG inputs, enabling diverse applications such as BCI, biomarker discovery, and real-time mental state monitoring.

The Methods section is organized as follows: First, the datasets employed in this study are introduced, including an EEG-fMRI dataset from a naturalistic viewing task, an EEG-fNIRS dataset from a motor imagery task, and additional datasets for clinical validation. Next, preprocessing steps for EEG, fMRI, and fNIRS are detailed to ensure cross-modal compatibility. Finally, the proposed framework's implementation is elaborated, encompassing the hyperdimensional integration strategy, feature extraction models, generative AI model, training protocols, and workflows. This framework provides a comprehensive exposition of the framework, from data preparation to its operationalization and application.

\subsection*{Data Description}
For the EEG-fMRI unified representation and cross-modal reconstruction experiments, an open-access dataset from a naturalistic viewing task with simultaneous EEG and fMRI recordings from 22 healthy adults (aged 23--51 years) was utilized \cite{dataset_1}. This dataset includes multiple visual stimulus conditions, with tasks assigned based on the films viewed. Specifically, Day 1 recordings—captured during the viewing of Despicable Me, Inscapes, Monkey1, and The Present—were labeled as Task 5, Task 6, Task 7, and Task 8, respectively. Day 2 recordings, obtained during the viewing of Despicable Me, Inscapes, Monkey2, and Monkey5, were designated as Task 1, Task 2, Task 3, and Task 4, respectively. For within-subject analysis, each participants Day 1 data served as the training set, with their corresponding Day 2 data (Tasks 1–4) as the test set. For cross-subject analysis, Day 1 data from participants 1–13 and 18–22 were used for training, while Day 1 data from participants 14–17 (renumbered as subjects 1-3 in the results presentation, Tasks 5–8) constituted the test set. The NODDI dataset \cite{noddi1_deligianni2016noddi, noddi2_deligianni2014relating}, comprising 64-channel EEG and whole-brain fMRI recordings from 17 healthy adults (11 males, 6 females; mean age 32.84 $\pm$ 8.13 years), was employed to train a model for generating BOLD from resting-state EEG. Here, data from 15 subjects (subjects 1–15) formed the training set, with data from subjects 16 and 17 reserved for testing. To assess clinical applicability in medical decision support, we utilized the EEG dataset of PD \cite{Parkinson_cavanagh2021eeg}, which includes 64-channel resting-state EEG recordings from 8 patients (4 males, 4 females; mean age 74.25 $\pm$ 8.75 years) and a healthy control group, to validate the NODDI-trained model. Given the small sample size and inter-subject variability in the PD dataset, data from four representative subjects (selected by age and gender) were used as the test set. As the PD dataset lacks paired fMRI data, the NODDI dataset—providing simultaneous EEG and fMRI recordings—was essential for training the EEG-to-BOLD mapping model.

The EEG-fNIRS experiments leveraged a motor imagery task dataset with simultaneous EEG and fNIRS recordings from 29 healthy participants \cite{dataset_2}. Data from participants 1–25 allocated to the training set and data from participants 26–29 to the test set.

\subsection*{Preprocessing}
EEG data were filtered using a bandpass filter with a range of 1–100 Hz to eliminate low-frequency drift and high-frequency noise. Artifacts related to eye movements and heart rate were subsequently removed to isolate brain-related signals.

For fMRI data, cortical surface mesh representations were reconstructed from each participant's structural images and registered to a common spherical coordinate system. Structural and functional images were aligned using boundary-based registration implemented in FreeSurfer \cite{freesurfer}. Resting-state fMRI data were mapped to this common spherical coordinate system by sampling from the mid-cortical ribbon in a single interpolation step \cite{nature4}. A 6-mm full-width half-maximum (FWHM) smoothing kernel was applied to the fMRI data in surface space, followed by downsampling to a mesh of 2562 vertices per hemisphere using the mri\_surf2surf function in FreeSurfer.

For fNIRS data, a 6th order zero-phase Butterworth bandpass filter with a passband of 0.01 to 0.1 Hz was applied after converting raw signals to deoxy-hemoglobin (HbR) and oxy-hemoglobin (HbO) data. HbR and HbO signals, collectively referred to as fNIRS, were uniformly processed throughout the experiment.

To account for the physiological delay between electrical brain activity and BOLD signals--typically a 6-second lag \cite{six_seconds}--data were segmented and temporally aligned accordingly. Specifically, the onset of blood oxygen signals was shifted to occur 6 seconds after the corresponding electrical activity, enabling the model to accurately capture the temporal relationship between EEG and BOLD responses for precise cross-modal representation.

\subsection*{Models}

\subsubsection*{Hyperdimensional Integration Strategy}

From the perspective of multiresolution cross-modal fusion, upper part of \cref{fig1}(b) illustrates the hyperdimensional integration strategy, which maps data from two distinct brain activity modalities into the spatio-temporal conjugate domain, achieving precise alignment across spatial and temporal dimensions. This mapping is guided by the spatial distribution and temporal dynamics of the data, constrained by their underlying phsysical relationships, thereby ensuring that the alignment retains anatomical and physiological relevance. The high temporal resolution of one modality complements the high spatial precision of the other, yielding a synergistic fusion where data are projected into a compatible representation within the spatio-temporal conjugate domain.

The high temporal resolution of one modality complements the high spatial precision of the other, enabling a synergistic fusion that projects data into the conjugate domain. By establishing a unified framework for multimodal brain function analysis, this approach facilitates the investigation of interregional coordination mechanisms and spatiotemporal interaction patterns.

To achieve hyperdimensional integration, the indirectly detected signals are assumed to be sampled at spatial points indexed by $i$, corresponding to sampling positions $p_i \in \mathbb{R}^3$. The directly detected signals are aligned with $p_i$ via a weighted average, with weights inversely proportional to the square of the Euclidean distance between $p_i$ and the directly detected sampling position $j$. The spatially aligned signal value $\hat{e}_i$ is defined as:

\begin{equation}
\hat{e}_i = \sum_{j} w_{ij} \cdot e_j,
\end{equation}
where $e_j$ represents the signal value at sampling point $j$ of the directly detected modality, and $w_{ij}$ is the weight calculated as:

\begin{equation}
w_{ij} = \frac{1}{\| p_i - r_j \|_2^2 + \epsilon},
\end{equation}
where $r_j \in \mathbb{R}^3$ denotes the coordinates of the directly detected sampling point $j$, $\| p_i - r_j \|_2$ is the Euclidean distance, and $\epsilon$ is a regularization term (e.g., $10^{-6}$) to avoid division by zero.

To account for temporal dynamic coupling, a time-alignment matrix $\mathbf{T} \in \mathbb{R}^{N \times N}$ is introduced, where the element $T_{kl}$ denotes the correlation between time points $k$ and $l$, defined as:

\begin{equation}
T_{kl} = \exp\left(-\frac{(t_k - t_l - \tau)^2}{2\sigma^2}\right),
\end{equation}
where $t_k$ and $t_l$ are the timestamps of the two modalities, respectively, $\tau$ is the physiological delay (approximately 6 seconds), and $\sigma$ is the Gaussian kernel width (e.g., 2 seconds) to reflect the distribution of temporal lag. The matrix $\mathbf{T}$ is optimized to achieve nonlinear alignment of cross-modal temporal dynamics.

Through spatial and temporal alignment, the signals are mapped onto a unified spatio-temporal template, ensuring anatomical consistency. This enhances the ability of subsequent modeling to learn the latent dynamic relationships, while preserving the complementary spatio-temporal characteristics of the signals, thereby providing a consistent representational framework for multimodal analysis.

\subsubsection*{Pre-trained Modal Feature Extraction Model}

The pre-trained feature extraction model is designed to process multimodal brain signals, including EEG and fMRI, into robust feature representations. For EEG data, the feature extraction model leverages a pre-trained architecture inspired by LaBraM-Base \cite{LaBraM}, a transformer-based model optimized for brain signal analysis. By incorporating parameters and module from LaBraM-Base, which was pre-trained on large-scale EEG datasets to capture hierarchical temporal and spatial patterns, this module ensures effective feature initialization adaptable to downstream tasks.

For fMRI data, a Transformer-based network is utilized for feature extraction. In temporal and spatial consistency experiments, this network was pre-trained on the UK Biobank dataset \cite{UKB_Sudlow2015UKBA}, which encompasses resting-state and task-based fMRI data from a large cohort. This pre-training captures the intricate spatial and temporal dependencies inherent in fMRI signals, yielding high-dimensional feature representations optimized for spatial analysis. By contrast, for downstream task evaluations and other experiments, a lightweight feature extraction layer with fewer parameters was adopted in place of the UK Biobank pre-trained model.

\subsubsection*{Generative AI Based Unified Representation Module}

The proposed framework in \cref{fig1}(b), serves as a cornerstone of the unified representation framework for integrating multimodal brain signals. Heterogeneous brain signals—such as EEG, fMRI, fNIRS, ECoG, and fPAI—are transformed into a cohesive representation through a generative learning framework. By leveraging generative AI techniques, this module enables cross-modal signal reconstruction, facilitating the cost-effective generation of functional neuroimages across modalities. The resulting unified representation captures both intra-modal and cross-modal dependencies, establishing a robust foundation for downstream applications, including signal denoising, modality translation, and neuroimage generation.

The proposed framework adopts the DiT model to implement this module, given its suitability for modeling the complex, high-dimensional, and heterogeneous nature of multimodal brain signals. Diffusion models are characterized by a forward diffusion process, during which noise is incrementally added to the data $ x_0 $ over $ T $ time steps:
\begin{equation}
q(x_t | x_{t-1}) = \mathcal{N}(x_t; \sqrt{1-\beta_t} x_{t-1}, \beta_t I),
\end{equation}
followed by a reverse denoising process, which is parameterized by the model:
\begin{equation}
p_\theta(x_{t-1} | x_t) = \mathcal{N}(x_{t-1}; \mu_\theta(x_t, t), \sigma_\theta(x_t, t)),
\end{equation}
where $ \beta_t $ represents the noise schedule, and $ \mu_\theta $ and $ \Sigma_\theta $ are learned parameters. This iterative denoising mechanism enables the progressive refinement of noisy representations into high-fidelity signals, making the approach well-suited for handling the temporal and modal variability inherent in brain activity data (e.g., neural electrical activity from EEG and blood oxygenation signals from fNIRS/fMRI).

The choice of a diffusion model over alternative generative frameworks, such as Generative Adversarial Networks (GANs) or Variational Autoencoders (VAEs), is driven by its superior generative fidelity and training stability. Unlike GANs, which are susceptible to mode collapse and training instability, or VAEs, which often struggle to model high-dimensional dependencies, diffusion models excel at capturing complex distributions through a progressive denoising process. This approach mitigates challenges such as overfitting and ensures consistent performance across diverse noise levels and sampling rates. The DiT architecture enhances this capability by replacing the traditional U-Net with Transformer blocks, incorporating self-attention and cross-attention mechanisms to effectively model intra-modal and cross-modal relationships within the data.

\subsubsection*{Workflows}

The workflow of the proposed Unified Representation Framework for low-cost cross-modality generation is illustrated in \cref{fig1}(c). This framework is designed to enable the transformation of low-cost, easily accessible modalities (e.g., EEG) into high-cost, highly restricted modalities (e.g., fMRI, fNIRS) through a unified representation learning framework. The process is structured to integrate multimodal brain signals into a cohesive representation, facilitating cross-modal generation while preserving both intra-modal and cross-modal dependencies.

The workflow is composed of several key stages, each formalized through mathematical representations for clarity. Initially, raw brain signals from different modalities---such as EEG, fMRI, or fNIRS---are processed through the hyperdimensional integration strategy, as detailed in the corresponding subsection, to ensure consistency across spatial and temporal dimensions. Features are then extracted using the Pre-trained Feature Extraction Model. For a given modality $ m $, the feature extraction process can be expressed as:
\begin{equation}
\mathbf{F}_m = \phi_m(\mathbf{X}_m; \theta_m),
\end{equation}
where $ \mathbf{X}_m $ represents the raw signal of modality $ m $, $ \phi_m(\cdot; \theta_m) $ denotes the pre-trained feature extractor (e.g., LaBraM for EEG data) with parameters $ \theta_m $, and $ \mathbf{F}_m $ is the extracted feature representation. This step ensures that modality-specific characteristics are effectively captured while enabling the framework to accommodate various types of functional signals.

The extracted features from all modalities are then fed into the Generative AI Based Unified Representation Module, which serves as the core of the proposed framework. This module maps the modality-specific features into a unified representation space, capturing both intra-modal and cross-modal dependencies. The unified representation $ x_0 $ is generated through a generative mapping function $ \mathcal{G} $, which can be formalized as:
\begin{equation}
x_0 = \mathcal{G}(\{\mathbf{F}_m\}_{m \in \mathcal{M}}; \theta),
\end{equation}
where $ \{\mathbf{F}_m\}_{m \in \mathcal{M}} $ denotes the set of features from all modalities $ \mathcal{M} $, $ \mathcal{G}(\cdot; \theta) $ represents the generative model (e.g., the DiT in this study) with parameters $ \theta $, and $ x_0 $ is the unified representation in a shared latent space. In this study, a DiT model is adopted within this module to facilitate the generation process, as detailed in the previous subsection.

To enable effective cross-modal generation, the generative model $ \mathcal{G} $ is trained using paired multimodal data to learn the relationships between modalities. During the training phase, paired data $ \{(\mathbf{X}_m, \mathbf{X}_{m'})\}_{m, m' \in \mathcal{M}} $ from source modality $ m $ (e.g., EEG) and target modality $ m' $ (e.g., fMRI) are utilized to optimize the model parameters $ \theta $. The training objective can be expressed as minimizing a reconstruction loss:
\begin{equation}
\mathcal{L}(\theta) = \mathbb{E}_{(\mathbf{X}_m, \mathbf{X}_{m'}) \sim \mathcal{D}} \left[ \ell(\mathbf{X}_{m'}, \hat{\mathbf{X}}_{m'}) \right],
\end{equation}
where $ \mathcal{D} $ represents the paired data distribution, $ \hat{\mathbf{X}}_{m'} $ is the reconstructed target modality signal, and $ \ell(\cdot, \cdot) $ is a loss function (e.g., mean squared error). This training process ensures that the generative model learns the cross-modal mapping, enabling the unified representation $ x_0 $ to encode modality-invariant features.

In the generation phase, the trained model is applied to perform cross-modal reconstruction. The unified representation $ x_0 $ is processed by a Modal Unpatcher to reconstruct the target modality signal. For a target modality $ m' $, the cross-modal generation process can be expressed as:
\begin{equation}
\hat{\mathbf{X}}_{m'} = \mathcal{D}_{m'}(x_0; \omega_{m'}),
\end{equation}
where $ \mathcal{D}_{m'}(\cdot; \omega_{m'}) $ is the decoding function (Modal Unpatcher) for modality $ m' $ with parameters $ \omega_{m'} $, and $ \hat{\mathbf{X}}_{m'} $ is the reconstructed signal in the target modality (e.g., fMRI or fNIRS). This step enables the generation of high-cost modality signals from low-cost inputs, such as reconstructing fMRI or fNIRS from EEG.

The proposed framework provides a robust framework for cross-modal generation, opening new avenues for advancing functional neuroimaging research. By enabling the generation of high-cost modality signals (e.g., fMRI, fNIRS) from low-cost inputs (e.g., EEG), this framework has the potential to significantly broaden the accessibility of advanced neuroimaging techniques, particularly in resource-constrained clinical and research settings. For instance, it can support the development of cost-effective diagnostic tools for neurological disorders by generating high-fidelity neuroimages without the need for expensive equipment. Meanwhile, the framework facilitates large-scale multimodal brain studies, fostering the discovery of novel biomarkers and enhancing the understanding of brain connectivity and dynamics.

\backmatter

\setcounter{figure}{0}


\noindent

\bibliography{sn-bibliography}

\renewcommand{\figurename}{Extended Data}
\renewcommand{\thefigure}{Fig.~\arabic{figure}}

\begin{figure*}[ht]
    \centering
    \includegraphics[width=0.8\textwidth]{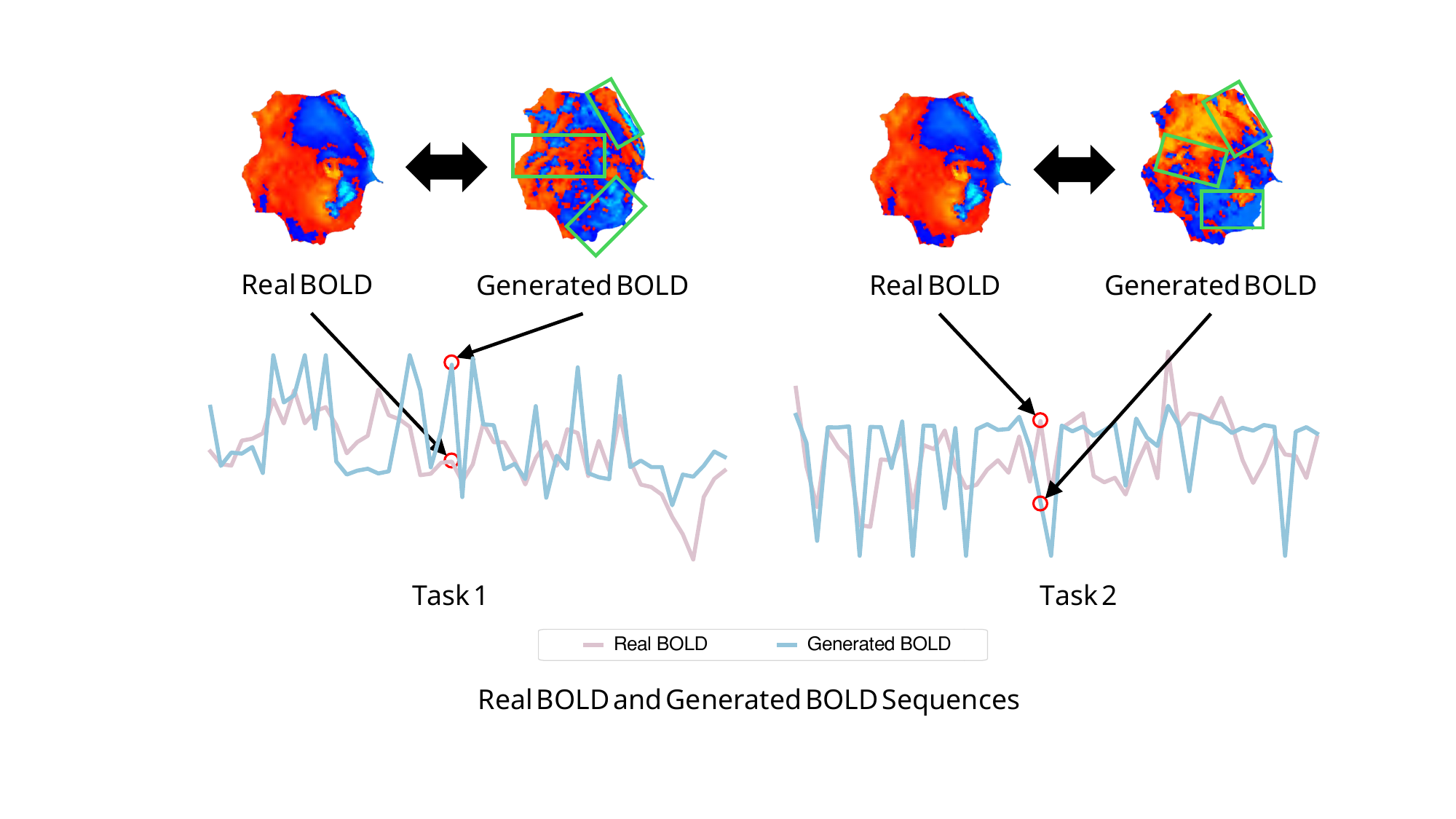}
    \caption{\textbf{Failure Case of the Proposed Framework}: The green boxes highlight shared key brain regions where the generated signals exhibit significant deviations from the real BOLD.}
    \label{fig:extended}
\end{figure*}

\end{document}